\documentclass[a4paper,french]{rnti}
\titrecourt{}

\nomcourt{H. Rezgui et M. Sassi-Hidri}

\titre{Modèle flou d'expression des préférences basé sur les CP-Nets}

\auteur{Hanene Rezgui\affil{1},
        Minyar Sassi-Hidri\affil{2}\
}

\affiliation{
    Université Tunis El Manar, Ecole Nationale d'Ingénieurs de Tunis\\BP. 37 le Belvédère 1002 Tunis, Tunisia\\
          \{\affil{1}hanene.rezgui, \affil{2}minyar.sassi\}@enit.rnu.tn\\

     }

\resume{%

This article addresses the problem of expressing preferences in flexible queries while basing on a combination of the fuzzy logic theory and Conditional Preference Networks or CP-Nets. }

\summary{%
This article addresses preferences expression in flexible queries by providing a fuzzy and explicit framework based on the combination of fuzzy logic and CP-Nets.
}
\begin{document}
\section{Introduction}
Les CP-Nets de \citet{boutilier99} ont constitué un formalisme pour la représentation graphique, intuitive et efficace des préférences utilisateur. En se basant sur ce formalisme, \citet{fatBou08} a proposé un modèle d'expression de ces dernières pour la recherche d'information (RI). Il autorise une quantification des données par des valeurs d'utilité définissant un ordre de préférence des instances de ces dernières. Cependant, l'affectation de ces valeurs d'utilité aux critères de la requête constituent une imprécision lors de sa réécriture et sa transformation en un CP-Net. Cette défaillance nous a induit au recours au formalisme de la logique floue dans la réécriture de la requête et sa transformation dans un CP-Net flou.
\section{Modèle flou d'expression des préférences}
Le modèle proposé bénéficie de la modélisation vague offerte par la logique floue et la représentation graphique et intuitive offerte par les CP-Nets. La première phase, pré-traitement, consiste à générer les bases de connaissances (BC) à partir de la base de données (BD) par une segmentation des données en régions floues. La deuxième phase, post traitement, formule les préférences utilisateur suite à une requête par le formalisme CP-Net et pondère, par la suite, cette dernière par le formalisme UCP-Net (Utility CP-Nets) \cite{bout01}.

Un CP-Net est vu comme un graphe orienté acyclique $G=(V,E)$ , où $V$ est un ensemble de noeuds $\{X_{1},...,X_{n}\}$ définissant les variables de préférence et $E$ un ensemble d'arcs orientés entre les noeuds qui traduisent des relations de dépendances préférentielles entre ces derniers. Toute variable $X_{i}$ est instanciable dans un domaine de valeurs $Dom(X_{i})=\{x_{i1},x_{i2},... \}$.

Pour modéliser les préférences d'un utilisateur, nous traduisons les variables de préférence de la requête par $Q_{num}=\bigcup_{t<T}(x_{1t}\max(\mu_{1j})_{t}\bigcap_{2 \leq i\leq N}(x_{it}\max(\mu_{ij})_{t}\bigcup_{k\neq t}x_{ik}\min(\mu_{ij})_{k}))$ où $t$ et $k$ sont les indices de valeur de la variable de préférence de l'ensemble $T$, $\mu_{ij}$ est le degré d'appartenance de la donnée $x_{i}$ au $j$\textit{ème} cluster (résultat d'une segmentation floue des données). Ensuite, pour formuler sa requête, l'utilisateur doit spécifier un ensemble de variables de ses préférences. Pour chaque variable donnée, l'utilisateur doit spécifier toutes ses dépendances préférentielles ainsi que l'ordre de préférences correspondants sur $Dom(x_{i})$. Le processus de pondération de la requête CP-Net coïncide avec la génération de l'UCP-Net. Il se base sur la propriété de dominance en utilisant les degrés d'appartenance $\mu_{ij}$.

Soit$x_{i}$ un noeud de la requête CP-Net tel que $Dom(x_{i})=k$, et soit $(\mu_{ij})_{t}$ le degré d'appartenance d'ordre $t$. Nous générons les utilités du noeud interne $x_{i}$ par $(\mu_{ij})_{1}=0,~(\mu_{ij})_{t}=(\mu_{ij})_{t-1}+S,~\forall~1<t<k$ puis nous calculons $Minspane(x_{i})=|(\mu_{ij})_{t+1}-(\mu_{ij})_{t}|$ et $Maxspane(x_{i})=|(\mu_{ij})_{k}-(\mu_{ij})_{1}|$.
\section{\'Evaluation}
Le but de l'évaluation est de calculer les degrés de pertinence des données fournies par la requête.

Chaque donnée pour une requête $Q(V,E)=\vee_{t}(\wedge_{i}(x_{it},f_{xit}))$ est représentée par un CP-Net. $D(V,E^{'})=\vee_{t}(\wedge_{i}(x_{it},(\mu_{ij})_{t})$ est représenté dans le même espace des critères que la requête. Nous réalisons ainsi  une projection des données de la BD sur l'espace de la requête. Nous attribuons une valeur d'importance de position $G_{X}$ à chacun des noeuds $x_{i}$ du CP-Net requête selon leurs niveaux respectifs dans le graphe. Pour tout noeud $x_{i}$, $G_{X}=1$ et pour tout noeud interne $x_{i}$, on note $B$ les descendants de $X$ et $G_{B}$ leurs ordres d'importance respectives. Nous avons alors $G_{X}=\max G_{B}+1$. Le poids agrégé $S_{k}$ est alors donné par $S_{k}=(\sum{i}((\mu_{ij})_{t}\ast G_{X_{i}})/(\sum{i}G_{X_{i}})$ où $X_{i}$ est le noeud contenant la donnée $x_{it}$ de la  BD.

Pour l'évaluation de la requête, on note $U_{k}$ le poids d'importance de $T_{k}$ dans $Q$. $F(D,T_{k})=S_{k}$ le poids de $T_{k}$ dans la donnée retournée $D$. On note par $Eval_{T_{k}}=(F(D,T_{k}),U_{k})$ la fonction d'évaluation de $T_{k}$  pour la donnée $D$. Les différentes conjonctions pondérées $(T_{k},U_{k})$ étant liées par une disjonction, ce qui donne $Eval_{T_{k}}=(F(D,T_{k}),U_{k})=\min(S_{k},U_{k})$ et $Eval(D,Q)=\max_{k}(\min(S_{k},U_{k}))$.
\bibliographystyle{rnti}

\bibliography{biblioexemple}
\Fr
\end{document}